%% file: main.tex
\documentclass[10pt,conference]{IEEEtran}

\usepackage{cite}

\ifCLASSINFOpdf
   \usepackage[pdftex]{graphicx}
   \graphicspath{{figs/}}
   \DeclareGraphicsExtensions{.pdf,.jpeg,.png}
\else
   \usepackage[dvips]{graphicx}
   \graphicspath{{../figs/}}
   \DeclareGraphicsExtensions{.eps}
\fi

\usepackage[cmex10]{amsmath}
\usepackage{amsthm}

\usepackage{algorithmic}

\usepackage{array}

\ifCLASSOPTIONcompsoc
  \usepackage[caption=false,font=normalsize,labelfont=sf,textfont=sf]{subfig}
\else
  \usepackage[caption=false,font=footnotesize]{subfig}
\fi

\usepackage{url}

\newif\iffinal
\finalfalse

\iffinal
\else
\usepackage[switch]{lineno}
\fi

\begin{document}
\title{A Machine Learning Approach for DeepFake Detection}

\author{\IEEEauthorblockN{Gustavo Cunha Lacerda and Raimundo Claudio da Silva Vasconcelos}
\IEEEauthorblockA{Instituto Federal de Educação, Ciência e Tecnologia de Brasília - IFB\\
Taguatinga-DF, Brasil\\
Email: gustavocunhalacerda@gmail.com, raimundo.vasconcelos@ifb.edu.br}
}

\maketitle

\input{texts/abstract}
\input{texts/introduction}
\input{texts/related_works}

\input{texts/materials}

\input{texts/proposed_method}
\input{texts/results_and_discussion}
\input{texts/conclusion}

\bibliographystyle{IEEEtran}

\bibliography{bib_ex}

\end{document}

%% file: texts/abstract.tex
\begin{abstract}
With the spread of DeepFake techniques, this technology has become quite accessible and good enough that there is concern about its malicious use. Faced with this problem, detecting forged faces is of utmost importance to ensure security and avoid socio-political problems, both on a global and private scale. This paper presents a solution for the detection of DeepFakes using convolution neural networks and a dataset developed for this purpose - Celeb-DF. The results show that, with an overall accuracy of 95\% in the classification of these images, the proposed model is close to what exists in the state of the art with the possibility of adjustment for better results in the manipulation techniques that arise in the future.
\end{abstract}

%% file: texts/introduction.tex
\section{Introduction}
Disinformation and its sharing are considered a worldwide concern entitled as fake news, that can be understood as the dissemination of content that is not true and that was purposely generated in order to convince readers about the veracity of information \cite{xul2008line}. 
Digitally manipulated images or videos to spread fake news have emerged and this technique is known as DeepFake, which is based on machine learning that offers a wide variety of methods for face exchange and manipulation, including the use of computer vision, deep learning  and word "fake".

This manipulation technology has been widely used in the cinema, as in the case of the film "Fast and Furious 7", in which the face of an actor, who died, was replaced in the body of his brother for the recording of the last scene of the film \cite{fastAndFurious}. In addition, this type of generation has already been used in the creation of a controversy speeches in a documentary about a cook who also passed away \cite{anthonyBourdain}. DeepFakes have also become very popular in the internet, going viral with memes, videos of famous politicians and artists. Despite its harmless use in entertainment product cases, with the evolution of imaging technologies and processing power, DeepFake has been used by people for malicious purposes. As home computers have increased their processing power to the point where most of the images generated could be made in a amateur way, this technology has become mainstream.

To avoid the harm caused by the misuse of this technology to privacy and veracity of information, some of the major technology companies have started initiatives to combat DeepFakes. An example of this is the DeepFake Detection Challenge (DFDC) \cite{dolhansky2020deepfake} initiative, which is a program developed by Facebook to promote solutions in detecting and classifying possibly manipulated images. In addition to these bigtech solutions, there has been an increase in research related to the classification of face manipulation in images. According to a projection made with data from Dimensions \cite{dimensions} by the end of 2020, about 737 DeepFake related papers were expected. However, according to the same site, using the same research method, there were 1,333 DeepFake related papers by the end of the year, an increase of about 80\% compared to the projection.  

Current DeepFake detectors show good classification results, even though, at the same time, face manipulation techniques have also received constant updates and improvements. Looking at this, the present work suggests an updated approach, starting with Celeb-DF \cite{li2019celeb} dataset, recently created for the purpose of clustering DeepFake images. In addition to use convolutional neural network models which, if well configured and fed with enough data, can detect manipulated images and, with finetuning, can classify new generations of face manipulation software. 

The goal of this paper is to develop a computer vision algorithm for deepfake detection with the help of deep learning. This work is organized as follows: section \ref{sec:related} presents related articles, section \ref{sec:materials} presents materials and the proposed method is shown in section \ref{sec:method}. Results and discussions are in section \ref{sec:results} and conclusions are in section \ref{sec:conclusions}.

%% file: texts/related_works.tex
\section{Related Works}
\label{sec:related}
With increasing discussions about the misuse of DeepFakes, academic researchers have improved research on detecting face manipulation. In \cite{guera2018deepfake} G\"{u}erra and Delp suggests using ImageNetV3 \cite{szegedy2016rethinking} pre-trained with ImageNet \cite{deng2009imagenet} dataset for feature extraction and using a Long Short-Term Memory (LSTM) for analysis and classification of results. The authors analyse the formation of deepfakes, in which encoders are applied to resize, extracting features of a face, to exchange these features with another face. In the process of exchanging these features, the target face and the original are not always in the same light conditions and even file format, which makes it difficult for the DeepFake creation algorithms to generate a realistic image. These errors in the creation can be targeted by algorithms that aim to decrypt them. This technique yielded an overall accuracy of 97.1\% on a 80-frame video fragment, demonstrating the high accuracy and effectiveness of this technique. 

Another important paper related to DeepFake detection is the researh of Lima \textit{et al} \cite{de2020deepfake}, that suggest an approach to detect manipulations on artifacts present in AI-generated DeepFakes videos in the frames transitions. The researchers used Celeb-DF and some pre-trained convolutional networks with Kinetics dataset to make the model for predictions. After train, the model achieved an average of 98.26\% in videos. 

In \cite{guarnera2020deepfake} it is also suggested the detection of DeepFakes through analysis of convolutional traces generated in the creation of this type of image. This new technique consists of a method of analyzing the relationship of each pixel and its neighbors, finding the relationship of these neighborhoods using expectation maximization. Then, after analyzing the relationships, it is possible to do a classification using KNN, SVM and LDA to define whether the image is a DeepFake. In this research there was a comprehensive analysis of different kernels and datasets combined which resulted in a maximum accuracy of 99.31\% using a linear SVM.

%% file: texts/materials.tex
\section{Materials}
\label{sec:materials}
\subsection{Hardware and Software}
For this work we used a computer with Intel(R) Xeon(R) CPU E3-1270 v6 @ 3.80GHz coupled with NVidia Titan V video card 16 Gb of \textsc{vram} and 66 Gb of \textsc{ram} DDR4@2400Mhz was used. The computer uses Ubuntu 20.04.3 LTS operating system and the main tool for building the convolutional network model was Pytorch library.

\subsection{Celeb-DF v2 and MediaPipe}
\begin{figure}[htp]
\centering
\includegraphics[width=3in]{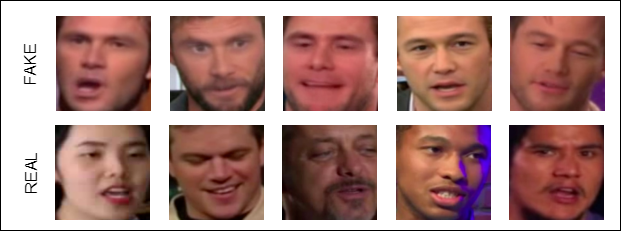}
\caption{CelebDF v2 dataset sample divided by class.}
\label{fig_dataset}
\end{figure}

The dataset chosen was Celeb-DF \cite{li2019celeb} in its second version. This dataset contains 590 videos without DeepFake, and 5,639 videos with DeepFake. The videos are about 13 seconds long at 30 frames per second, totaling over two million frames of data for use in DeepFake classification problems. In Figure \ref{fig_dataset}, it is possible to see a batch with examples of the two classes that is is the dataset. 

\begin{figure}[htp]
\centering
\includegraphics[width=2.5in]{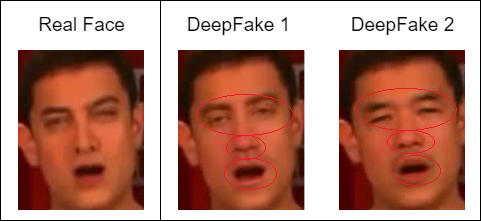}
\caption{Example of face critic regions and artifacts.}
\label{fig_fakexreal}
\end{figure}

In this image set the people's faces underwent several deepfaking techniques that resulted in manipulated videos to compose the class of fakes present in the set. It is noticeable that during the process certain regions end up being affected in the fusion between target face and original, forming image artifacts. The most critic regions of these errors are mouth, nose and eyes, which is exemplified in the Figure \ref{fig_fakexreal}.

Another tool that was be used in conjunction with the dataset is the computer vision toolkit developed by Google programmers and researchers - MediaPipe \cite{lugaresi2019mediapipe}. It is a set of computer vision tools created with a focus on efficiency and portability in which even embedded systems can run applications that consume the library. In this work, the set of tools related to point detection and face meshing in images was used.

%% file: texts/proposed_method.tex
\section{Proposed Method}
\label{sec:method}
\begin{figure}[htp]
\centering
\includegraphics[width=2.5in]{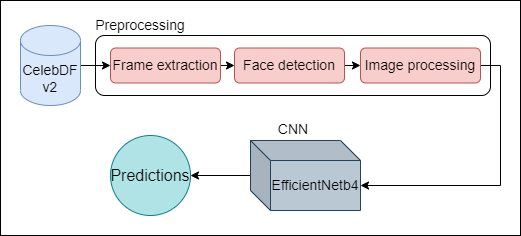}
\caption{Proposed method}
\label{fig_method}
\end{figure}

The method proposed here is composed of three main steps: pre-processing, training and validation, as we can see in Figure \ref{fig_method}. After choosing the dataset, videos are processed to extract frames and faces that will be used as input to the network for training step. These data are then normalized and their sizes adjusted for homogenization purposes. Training is the kernel of the research, in which a pre-trained convolutional neural network model was chosen to go through a fine-tuning phase to adjust network parameters and characteristics for the DeepFakes binary classification problem.

\subsection{Data processing}
To carry out the training with the Celeb-DF dataset, it was first necessary to balance the amount of videos of the two classes. The Celeb-DF in its second version had 590 real videos and 5639 fake videos, so the limit of videos for each class was set to 590, resulting in 1780 videos regarding the binary classification of the problem. Then the videos were separated into a test subset to ensure that in training subset there is no frame of the same video that is also in the test subset. This step ensures that the tests are blind and that the network has never trained with the data used for this purpose. Table \ref{table_videos_division} shows this separation.

\begin{table}[]
\renewcommand{\arraystretch}{1.3}
\caption{Subset videos division}
\label{table_videos_division}
\centering
\begin{tabular}{|c||c|}
\hline
\textbf{Fake Train Subset} & 472\\
\hline
\textbf{Real Train Subset} & 472\\
\hline
\textbf{Fake Test Subset} & 118\\
\hline
\textbf{Real Test Subset} & 118\\
\hline
\end{tabular}
\end{table}

\begin{figure}[htp]
\centering
\includegraphics[width=2.5in]{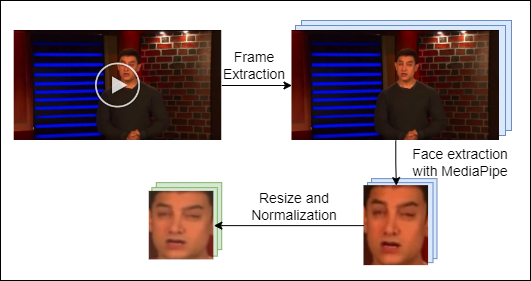}
\caption{Video processing steps}
\label{fig_img_processing}
\end{figure}

After this separation, up to 500 frames of each video were extracted, counting only frames in which the MediaPipe library was able to find a human face with an 80\% certainty rate. In addition, the facial points found by google's library were used to make a cut on the boundaries of the face, decreasing the information and focusing on what the network must learn to perform the classification. These faces went through normalization and resizing by 224x224 pixels before being saved in a dataset that will be used by the EfficientNet network. The details of the processing flow is showed in Figure \ref{fig_img_processing}.

\subsection{EfficientNet}
A Convolutional Neural Network (CNN) is a Machine Deep Learning algorithm that can capture an input image, assign weight and bias to various characteristics of an image, differentiate objects, and perform less expensive analysis on image sets \cite{cnns2022data}. The prepossessing required on a CNN is much lower compared to other ranking algorithms. While in primitive methods filters are handmade, with sufficient training, CNNs have the ability to learn these filters.


With the advancement and dissemination of the power of CNNs, architectures emerged that sought to extract the most from this concept, such as ResNet or Xception. In this research, the EfficientNet \cite{koonce2021efficientnet} architecture was chosen, a convolutional neural network model that is very efficient in relation to the amount of resources and interactions for convergence.

While EfficientNet work well in ImageNet \cite{deng2009imagenet}, it should also be transferable to other datasets to be as useful as possible. EfficientNet was tested on eight widely used transfer learning datasets. EfficientNet models achieve better accuracy with 4.7x average (up to 21x) parameter reduction in 5 of the 8 datasets with transfer learning compared to the state-of-art solutions. Such as CIFAR-100 \cite{balajicifar100} (91.7\%) and Flowers \cite{nilsback2008automated} (98.8\%) results of accuracies suggest that the architecture is highly recommended for problems that can be solved by transfer learning or fine tuning, which is the case for DeepFake classification.

\subsection{Train and validation}
This work used a pre-trained version of EfficientNet-B4, an architecture of the EfficientNet family, to train the model. This version was chosen because it presents the best cost-benefit ratio with the hardware available for the present research. A learning rate of 0.0001 and Adam optimizer with 0.005 weight decay used to avoid overfitting. The dataset for training was divided into the 80/20 training and validation ratio, respectively. The model was trained with dataset in batches of 32 faces in 80 epochs, it also performs a validation step in each epoch to verify the quality of predictions and the convergence of the classification.

%% file: texts/results_and_discussion.tex
\section{Results and Discussion}
\label{sec:results}
\begin{figure}[!t]
\centering
\includegraphics[width=2.5in]{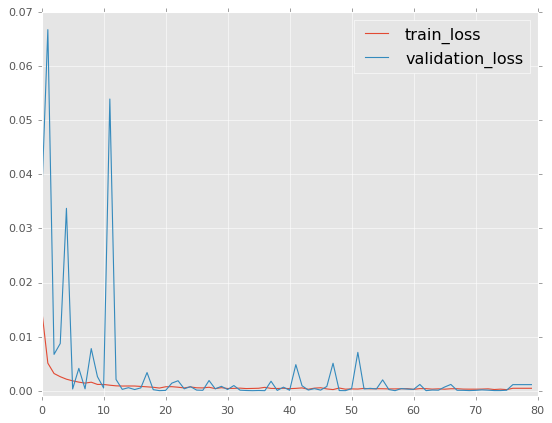}
\caption{Validation and train losses over the epochs.}
\label{fig_val_train_losses}
\end{figure}

An EfficientNet-B4 architecture and Celeb-DF V2 dataset presented very satisfactory results when used together. During training and validation, accuracy progression as shown in Figure \ref{fig_val_train_losses}, remained above 97\% at all times with increasing trends. The validation and training losses remained in a decreasing trend over the iterations, indicating the convergence of the model throughout the training stage.

\begin{figure}[!t]
\centering
\includegraphics[width=2.5in]{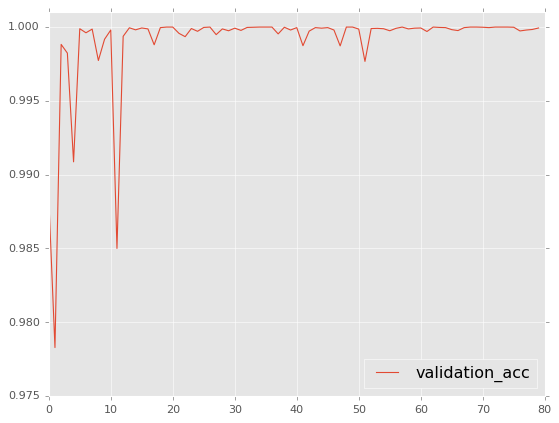}
\caption{Validation accuracy over the epochs.}
\label{fig_val_acc}
\end{figure}

Some metrics were used for the analysis of the proposed model, among them F1 Score, Accuracy, Precision and Recall with results presented in Table \ref{table_results}. 

\begin{table}[]
\renewcommand{\arraystretch}{1.3}
\caption{Final results of tests with the model}
\label{table_results}
\centering
\begin{tabular}{|c||c|}
\hline
\textbf{Final Accuracy} & 0.9552\\
\hline
\textbf{Recall} & 0.9161\\
\hline
\textbf{Precision} & 0.9999\\
\hline
\textbf{F1 Score} & 0.9562\\
\hline
\end{tabular}
\end{table}

\begin{figure}[htp]
\centering
\includegraphics[width=2.5in]{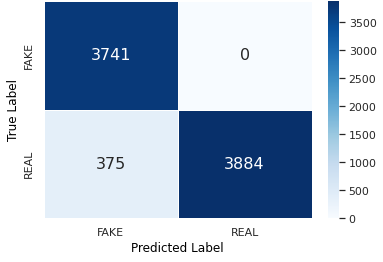}
\caption{Confusion matrix in the model applied on the images test subset.}
\label{fig_conf_matrix}
\end{figure}

Another important metric that was used to qualify the method generated by this paper is the confusion matrix. With this metric is possible to verify the performance of the algorithm by comparing the predictions with the real values of the labels. In our binary classification model there are 4 results that can be seen in the Figure \ref{fig_conf_matrix}: true positive (TP) true negative (TN), false positive (FP) and false negative (FN) where true or false is understood as whether an image is a DeepFake or not. In it, it can be seen that the model tends to be more rigorous in correctly classifying real images, which is good for this type of classification considering its applicability. There is more sensitivity in classifying an image with any suspicion of manipulation, in order to ensure safety rather than certainty that the image is true fake. This behavior also explains why the precision metric resulted in a value close to 1, since there is a only one real prediction in fake image.

Although the final accuracy was below the state-of-the-art of 99.73\% seen in \cite{guera2018deepfake} or 97.1\% seen in \cite{guarnera2020deepfake}, the results are as good as there is the possibility of improvement and the network can still be fed with new data in the current state for finetuning and can adapt to new models of DeepFake creation. Besides, the Celeb-DF v2 dataset proved to be quite sophisticated compared to others used in other research on digitally manipulated face classification such as the DFDC \cite{dolhansky2020deepfake} datasets.

%% file: texts/conclusion.tex
\section{Conclusions}
\label{sec:conclusions}
DeepFake detection is something quite complex, since techniques for creating new manipulated images arrives faster than good solutions for detection and protection against this type of spoofing. Thus, the technique presented by this article presented very satisfactory results for the current state of the problem. The EfficientNet network in conjunction with the Celeb-DF dataset were assertive and combined for the generation of a robust model that achieved satisfactory predictions, maintaining an accuracy of more than 93\% in images of 224x224 pixels of height and width, achieving Recall of 0.9161 and F1 of 0.9562.

With the results found it is possible to find a way to improve the model. For future work it is necessary that the model presents tests on images from other sources and more varied manipulation techniques, increasing the generalization of the model for predictions. Moreover, increasing the number of iterations seems to be beneficial for the model, since the results showed a tendency to increase for accuracy and decrease for losses without reaching an overfitting that would lead to more divergent results in the tests. Another factor that can be decisive for improving classification performance is, with greater features, choosing a more complex CNN model or one of the other models in the EfficientNet family that has more parameters and a larger kernel.